\documentclass[10pt,twocolumn,letterpaper]{article}

\usepackage{iccv}
\usepackage{times}
\usepackage{epsfig}
\usepackage{graphicx}
\usepackage{amsmath}
\usepackage{amssymb}
\usepackage{microtype}      
\usepackage{multirow}
\usepackage[american]{babel}

\newcommand{\squishlist}{
	\begin{list}{$\bullet$}
		{ \setlength{\itemsep}{0pt}
			\setlength{\parsep}{1pt}
			\setlength{\topsep}{1pt}
			\setlength{\partopsep}{0pt}
			\setlength{\leftmargin}{1.5em}
			\setlength{\labelwidth}{1em}
			\setlength{\labelsep}{0.5em} } }
\newcommand{\squishend}{\end{list} 
}
\graphicspath{{figures/}}
\usepackage{subfig}

\usepackage[pagebackref=true,breaklinks=true,letterpaper=true,colorlinks,bookmarks=false]{hyperref}

\iccvfinalcopy 

\def\iccvPaperID{4216} 

\ificcvfinal\fi

\begin{document}

\title{NMS Threshold matters for Ego4D Moment Queries}

\author{Lin Sui$^{1}$, Fangzhou Mu$^{2}$, Yin Li$^{2}$\\
$^1$State Key Laboratory for Novel Software Technology, Nanjing University\\ $^2$University of Wisconsin-Madison\\
{\tt\small suilin0432@gmail.com, \ fmu2@wisc.edu, \ yin.li@wisc.edu}
}

\maketitle
\ificcvfinal\thispagestyle{empty}\fi

\begin{abstract}
This report describes our submission to the Ego4D Moment Queries Challenge 2023. Our submission extends ActionFormer~\cite{zhang2022actionformer}, a latest method for temporal action localization. Our extension combines an improved ground-truth assignment strategy during training and a refined version of SoftNMS at inference time. Our solution is ranked 2$nd$ on the public leaderboard with 26.62\% average mAP and 45.69\% Recall@1x at tIoU=0.5 on the test set, significantly outperforming the strong baseline from 2023 challenge. Our code is available at \url{ https://github.com/happyharrycn/actionformer\_release}.
\end{abstract}

\section{Introduction}

The Ego4D Moment Queries (MQ) task aims to localize all moments of actions in time and recognize their categories within an untrimmed egocentric video. We adopt a two-stage approach for this task, where clip-level features are first extracted from raw video frames using a pre-trained feature network, followed by a temporal localization model that predicts the onset and offset of action instances as well as their categories. Our submission last year explored the combination of a latest localization model (ActionFormer~\cite{zhang2022actionformer}) and a strong set of video features~\cite{mu2022strong}. This work seeks to improve the localization model .

A limitation of ActionFormer~\cite{zhang2022actionformer} lies in its label assignment at training time; annotated action instances are assigned to candidate moments based on \textit{center sampling}, a heuristic that designates positive labels to moments proximal to the center of an action instance. Recent literature in object detection, however,  shows that such static assignment strategy is insufficient for complex spatial configuration of objects.  Inspired by this insight, we propose to adapt SimOTA~\cite{ge2021yolox}, a dynamic label assignment strategy, for temporal action localization. SimOTA assigns ground-truth action instances to the candidate moments on the fly by solving an optimal transport problem. Further, we refine SoftNMS to account for densely overlapping actions in Ego4D.   

Equipped with these modifications, our solution extends our prior work and is ranked 2$nd$ on the public leaderboard. Specifically, our solution attains 26.62\% average mAP and 45.69\% Recall@1x at tIoU=0.5 on the test set, significantly outperforming the strong baseline from 2023 challenge. We hope our work will shed light on future development in temporal action localization and egocentric vision.

\section{Method}

\begin{figure*}[t!]
	\centering 
	\includegraphics[width=0.8\linewidth]{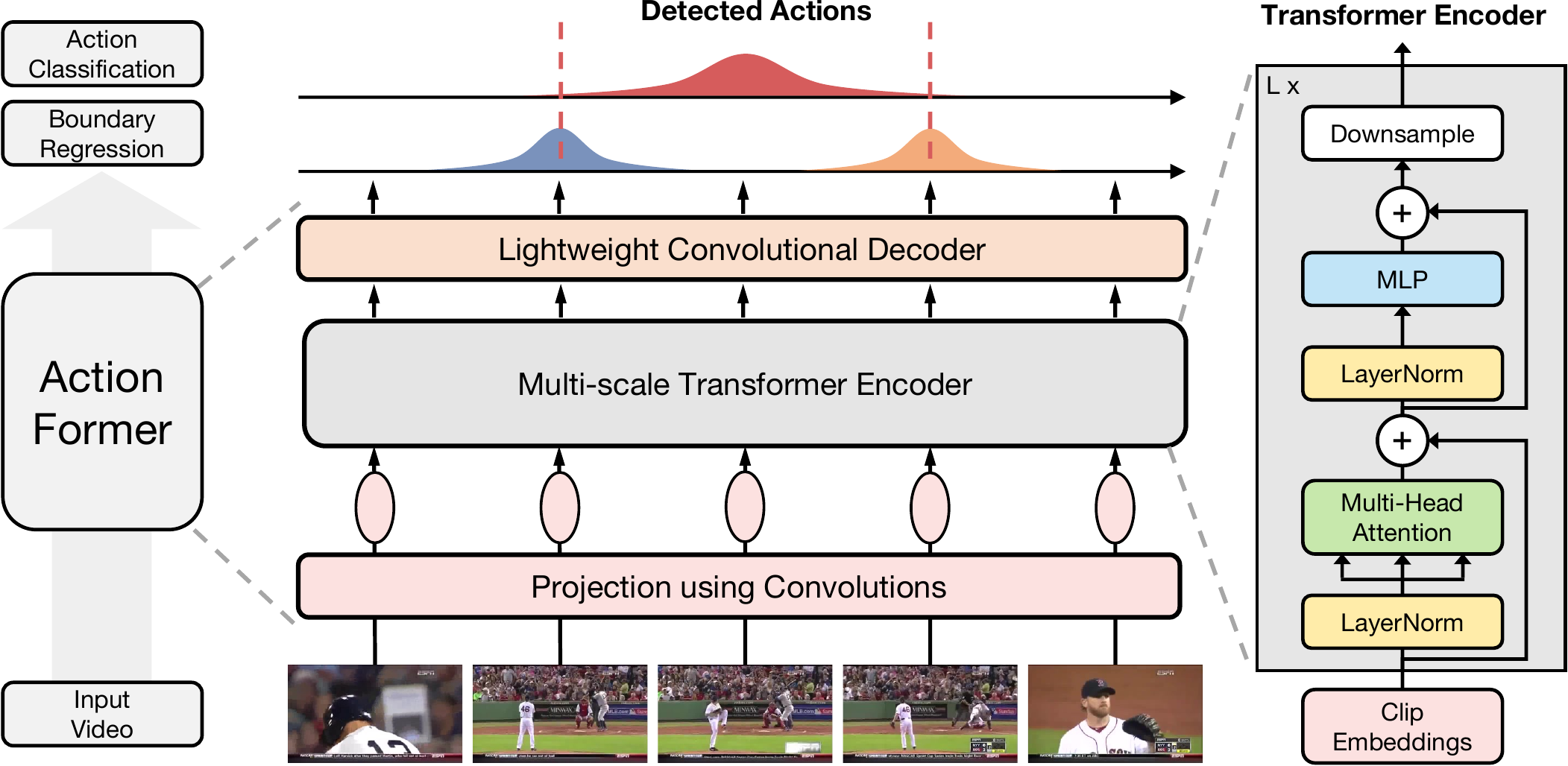}\vspace{-0.5em}
	\caption{\textbf{Overview of ActionFormer.} Taken from~\cite{zhang2022actionformer}.}
	\label{fig:overview}
\end{figure*}

Our method builds on ActionFormer~\cite{ego4d2022actionformer}, the state of the art for temporal action localization, yet introduces two key modifications. First, we adopt SimOTA~\cite{ge2021yolox}, a dynamic ground truth assignment strategy in training. Second, we flatten the Gaussian penalty function in SoftNMS~\cite{softnmsiccv2017} to account for ground truth moments with significant overlap. We now present the key components of our method.

\paragraph{Moment Localization with ActionFormer.} ActionFormer~\cite{zhang2022actionformer} takes as input a 1D sequence of clip-level video features and builds a feature pyramid using local self-attentions. This pyramid serves as a multi-scale representation of moment candidates. Each location on the pyramid defines the center of a moment, whose temporal scale is determined by the pyramid level (\emph{i.e.}, longer moments reside on higher levels of the pyramid). A classification head subsequently assigns a confidence score to each moment, whereas a regression head predicts the distances from the center of a moment to its onset and offset. These predictions are decoded into action segments and further combined using SoftNMS~\cite{softnmsiccv2017}. We refer readers to~\cite{zhang2022actionformer} for more details on ActionFormer.

\paragraph{Dynamic Label Assignment.} ActionFormer follows a \emph{fixed} set of rules collectively known as center sampling to convert ground-truth action segments into point-wise classification labels. Center sampling was first seen in the literature of single-stage object detection~\cite{tian2019fcos}, in which it has lately been superseded by more powerful, \emph{dynamic} label assignment strategies~\cite{ge2021ota,ge2021yolox} that evolve in tandem with training losses. In this work, we report a similar finding that training ActionFormer with SimOTA~\cite{ge2021yolox}, an efficient dynamic label assignment technique, yields a small yet significant performance gain compared to using center sampling. We provide ablation results in Section~\ref{experiments} and refer readers to~\cite{ge2021yolox} for more details on SimOTA.

\paragraph{Dealing with Near-Replicates.} Our initial exploratory analysis revealed that $15\%$ of ground-truth moments in the Ego4D-MQ dataset are \emph{near-replicates} (\emph{i.e.}, $\ge 90\%$ overlap with another moment). This presents a unique challenge to ActionFormer, which relies on aggressive non-maximum suppression (NMS) to reduce highly overlapping predictions, thereby harming precision at high recall in the presence of near-replicates. In this work, we propose to tune the standard deviation $\sigma$ of the Gaussian penalty function $f$ in SoftNMS~\cite{softnmsiccv2017} as a simple fix. Intuitively, a small $\sigma$ as recommended by~\cite{softnmsiccv2017} yields a peaky $f$ that incurs strong penalty on near-replicates, whereas $f$ flattens out as $\sigma$ increases, leaving near-replicates less affected. We empirically found that setting $\sigma$ to the unusually large value of 2 brought a significant improvement in mAP by 1.8 absolute percentage point compared to using the default value of 0.9 as in~\cite{ego4d2022actionformer}.

\begin{table*}[t!]
    \centering
    \begin{tabular}{c|c||c|c|c}
    \hline
       Split & Features & SimOTA & SoftNMS $\sigma$ & average mAP \\
       \hline 
       \multirow{6}{*}{Val} & SlowFast + Omnivore + EgoVLP & & 0.9 & 21.40 \\
       & InternVideo + Omnivore + EgoVLP & & 0.9 & 24.11 \\
       & InternVideo + Omnivore + EgoVLP & \checkmark & 0.9 & 24.41 \\
       & InternVideo + Omnivore + EgoVLP & \checkmark & 1.5 & 25.71 \\
       & InternVideo + Omnivore + EgoVLP & \checkmark & 2.0 & \textbf{26.07} \\
       & InternVideo + Omnivore + EgoVLP & \checkmark & 4.0 & 25.76 \\
       \hline 
       \multirow{2}{*}{Test} & InternVideo + Omnivore + EgoVLP & \checkmark & 0.9 & 25.33 \\
        & InternVideo + Omnivore + EgoVLP & \checkmark & 2.0 & \textbf{26.62} \\
         \hline 
    \end{tabular}
    \caption{\textbf{Results on Ego4D Moment Queries dataset.}}\label{table:results}
\end{table*}

\section{Experiments and Results}\label{experiments}

We now present our experiments and results.

\paragraph{Evaluation Protocol and Metrics.} We follow the official \emph{train}/\emph{val}/\emph{test} splits for evaluation. Our model is trained on the \emph{train} split when results are reported on the \emph{val} split, and is trained on the \emph{train} and \emph{val} splits combined when results are submitted for final evaluation on the \emph{test} split. In line with the official guideline, we adopt average mAP as our main evaluation metric.

\paragraph{Implementation Details.} We use pre-extracted Omnivore~\cite{girdhar2022omnivore}, EgoVLP~\cite{kevin2022egovlp} and InternVideo~\cite{chen2022internvideo} features from raw videos as input to ActionFormer, and set the embedding dimension throughout the model to 1152. Following the official code release, we train ActionFormer using the AdamW optimizer~\cite{loshchilov2017decoupled} for 15 epochs with a mini-batch size of 2, a learning rate of 0.0001 and a weight decay of 0.05.

\begin{figure*}[t!]
	\centering 
	\includegraphics[width=1.0\linewidth]{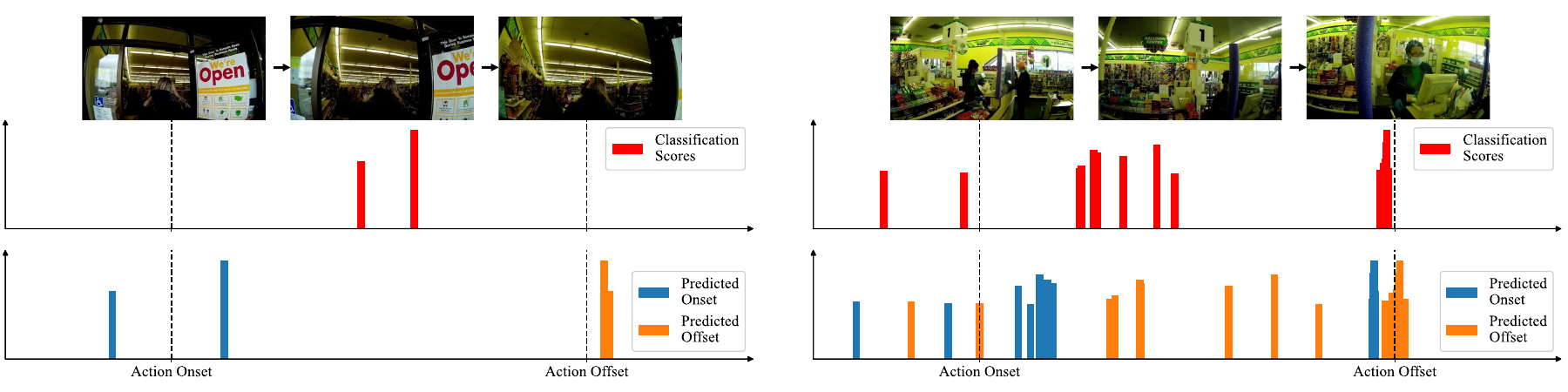}
	\caption{\textbf{Result Visualizations.} From \emph{top} to \emph{bottom}: (1) input video frames;  (2) action scores at each time step; (3) histogram of action onsets and offsets computed by weighting the regression outputs using action scores. \emph{Left}: a success case for ActionFormer. \emph{Right}: a failure case with multiple center regions and wrong onset/offset regression.}
	\label{fig:result_viz}
\end{figure*}

\begin{figure}[t!]
	\centering 
	\includegraphics[width=1.0\linewidth]{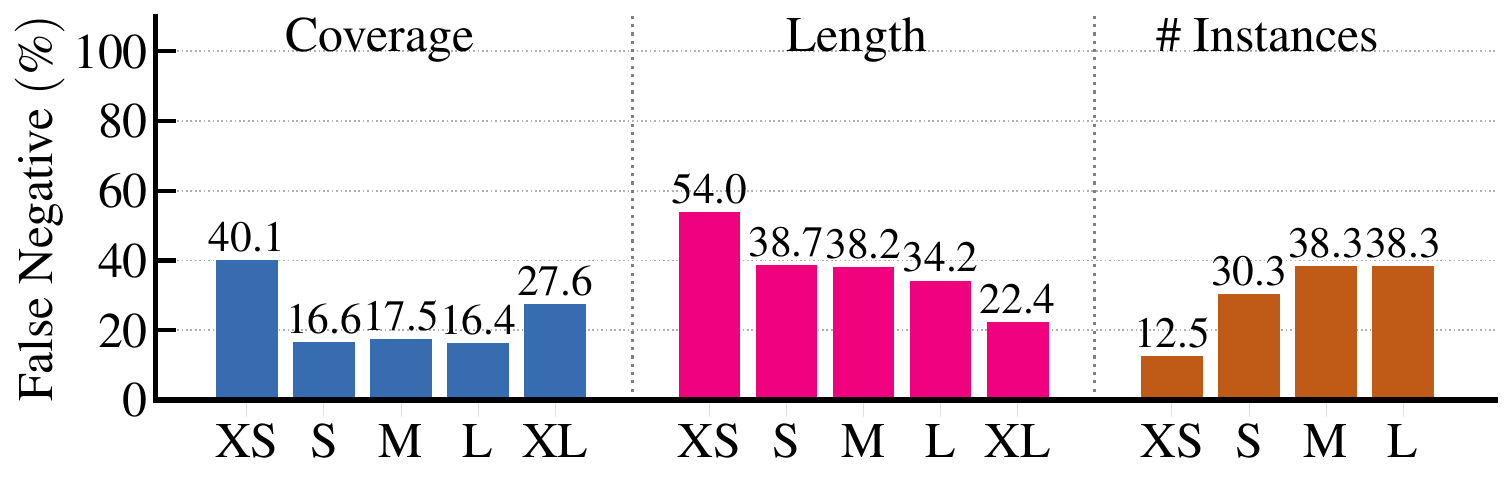}
	\caption{\textbf{False Negative (FN) Analysis with respect to different moment characteristics.} Our method tends to miss short actions as well as actions in videos with short moment coverage. It also exhibits higher FN rate on videos with multiple actions, possibly due to near-replicates.} 
	\label{fig:fn}
\end{figure}

\begin{figure}[t!]
	\centering 
	\includegraphics[width=1.0\linewidth]{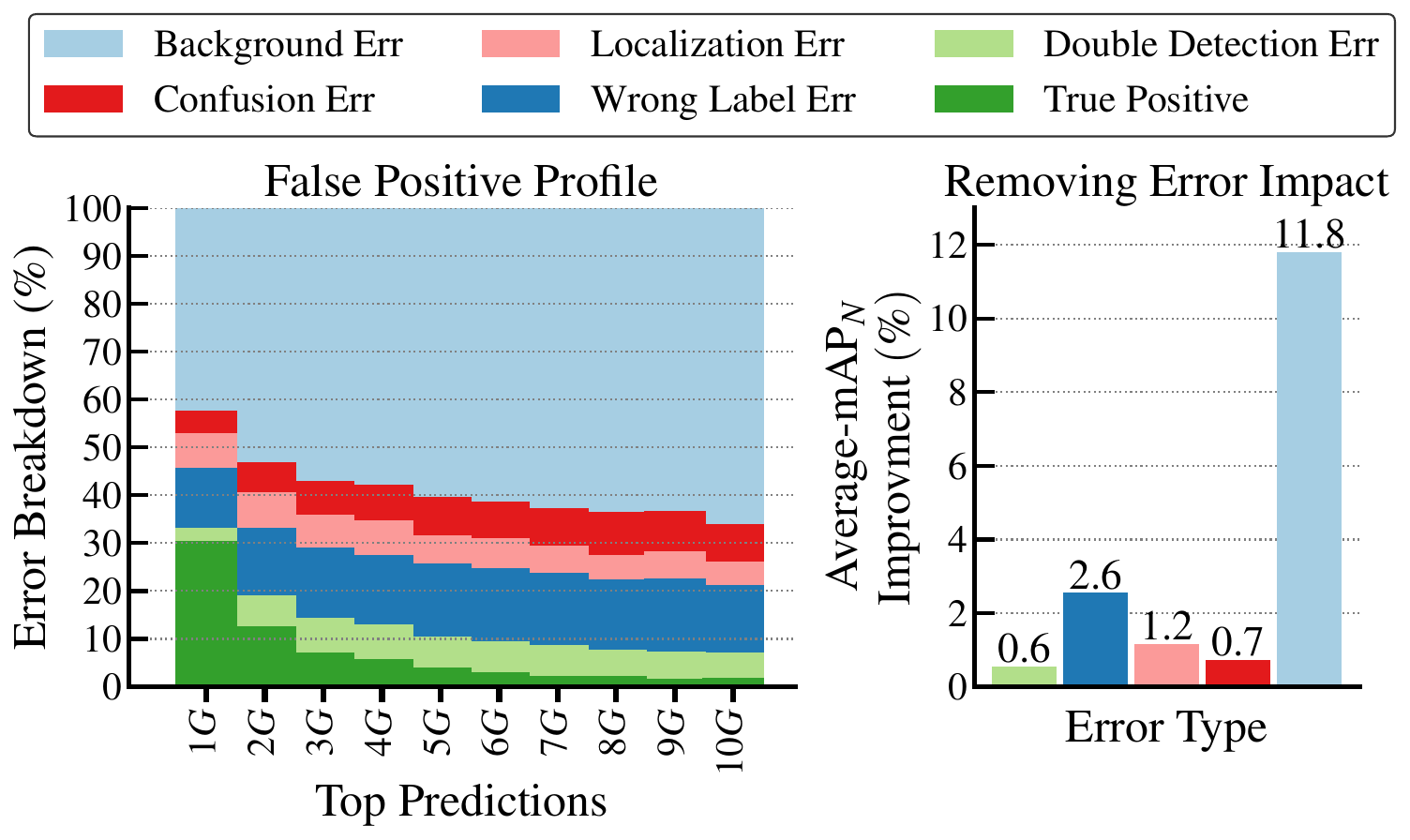}
	\caption{\textbf{False Positive (FP) Analysis using DETAD~\cite{alwassel2018diagnosing}.} \emph{Left}: FP error breakdown when considering the predictions for the top-10 ground-truth (G) instances. \emph{Right}: The impact of error types. Background error and wrong label error are the top two error types.} 
	\label{fig:fp}
\end{figure}

\begin{figure}[t!]
	\centering 
	\includegraphics[width=1.0\linewidth]{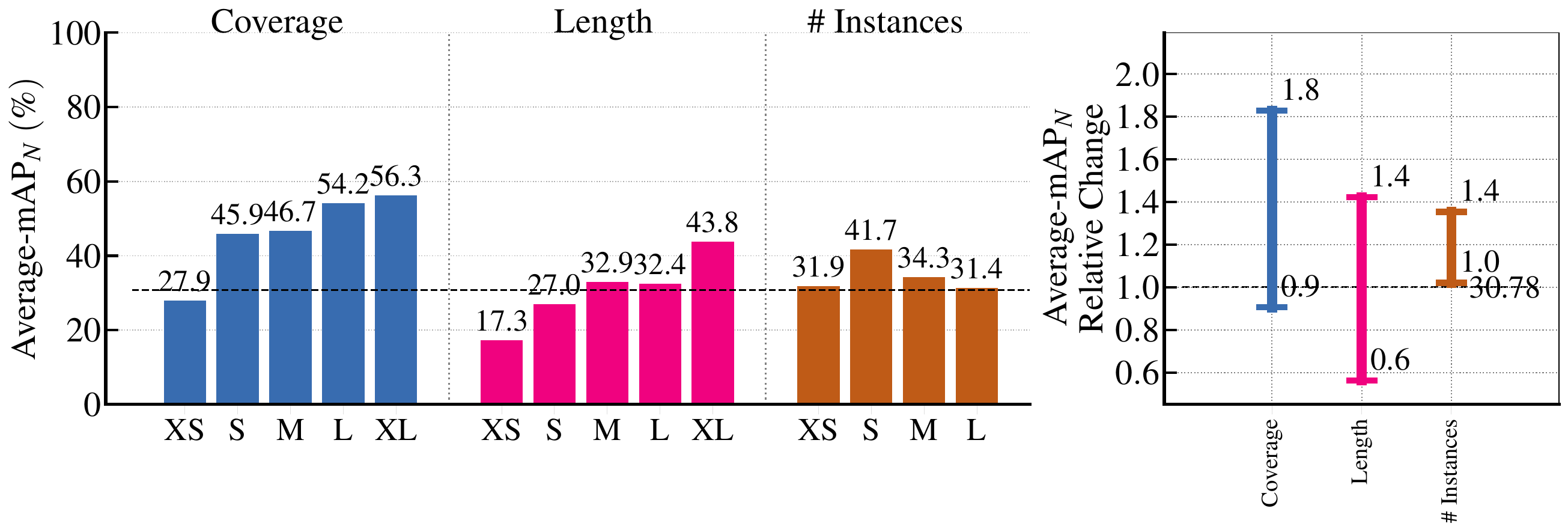}
	\caption{\textbf{Sensitivity Analysis with respect to different moment characteristics.} \emph{Left:} Normalized mAP at tIoU=0.5. Our method performs better on videos with high moment coverage and is more capable of detecting long actions. \emph{Right:} The relative normalized mAP change at tIoU=0.5. Performance of our method is most sensitive to moment coverage and duration.} 
	\label{fig:sensitivity}
\end{figure}

\paragraph{Results.} Our results on the \emph{val} and \emph{test} splits are summarized in Table~\ref{table:results}. Replacing the official SlowFast features with InternVideo features brings a notable 2.71 absolute-percentage-point improvement on average mAP on the \emph{val} split. This highlights the strength of InternVideo as a video foundation model for representation learning. The introduction of dynamic label assignment further boosts the average mAP by a small yet significant 0.3 absolute percentage point. Finally, the largest performance gain ($>$1.3 absolute percentage points on both \emph{val} and \emph{test} splits) is attained by tuning the spread of Gaussian penalty function in SoftNMS to account for near-replicates. With everything combined, ActionFormer reaches an average mAP of 26.07$\%$ on the \emph{val} split and 26.62$\%$ on the \emph{test} split. Figure~\ref{fig:result_viz} provides visualizations of model predictions. 

\paragraph{Limitations.} We present false negative analysis (Figure~\ref{fig:fn}), false positive analysis (Figure~\ref{fig:fp}) and sensitivity analysis (Figure~\ref{fig:sensitivity}) of our method. Our method demonstrates stronger performance on videos with high moment coverage and is more capable of identifying longer actions. It exhibits significantly higher error rate on videos with low moment coverage and was not able to accurately localize short actions. This is further manifested by the surprisingly high background error rate.

\section{Conclusion}
In this report, we described our solution to the Ego4D Moment Queries Challenge 2023. Our solution is based on ActionFormer, yet introduces two key modifications to training and post-processing that brought substantial performance gain without change to the model architecture. We provided extensive analysis of our results which highlights the strength and weakness of our approach. We hope our solution and results can offer new insights to the Ego4D MQ task.

\paragraph{Acknowledgement.} We thank Chen-Lin Zhang for fruitful discussions about ActionFormer.

{\small
\bibliographystyle{ieee_fullname}
\bibliography{egbib}
}

\end{document}